\documentclass[format=acmsmall, review=false]{acmart}

\usepackage{acm-ec-18}

\usepackage{booktabs} 

\usepackage[ruled]{algorithm2e} 

\SetAlFnt{\small}
\SetAlCapFnt{\small}
\SetAlCapNameFnt{\small}
\SetAlCapHSkip{0pt}
\IncMargin{-\parindent}

\usepackage{xcolor}
\usepackage{soul}
\usepackage[utf8]{inputenc}

\usepackage{float}
\usepackage{amssymb}
\usepackage[noend]{algpseudocode}
\usepackage{algorithmicx}
\usepackage{graphicx}
\usepackage{amsmath,amsthm,mathrsfs,bm}
\usepackage{amsfonts}  
\usepackage{amssymb}  
\usepackage{etoolbox}
\usepackage{array}
\usepackage[inline]{enumitem}
\usepackage{setspace}
\usepackage{subfigure}
\usepackage{tikz}  
\usepackage{multirow}
\usepackage{url}

\makeatletter
\newcommand{\algmargin}{\the\ALG@thistlm}
\makeatother
\newlength{\whilewidth}
\algdef{SE}[parWHILE]{parWhile}{EndparWhile}[1]
{\parbox[t]{\dimexpr\linewidth-\algmargin}{%
		\hangindent\whilewidth\strut\algorithmicwhile\ #1\ \algorithmicdo\strut}}{\algorithmicend\ \algorithmicwhile}%
\algnewcommand{\parState}[1]{\State%
	\parbox[t]{\dimexpr\linewidth-\algmargin}{\strut #1\strut}}

\newcommand{\ma}{\mathcal A}

\newcommand{\ml}{\mathcal L}

\newtheorem{thm}{Theorem}
\newtheorem{dfn}{Definition}

\newtheorem{ex}{Example}

\AtEndEnvironment{ex}{}
\let\oldeg\ex
\renewcommand{\ex}{\oldeg\normalfont}

\usepackage{array}
\newcolumntype{P}[1]{>{\centering\arraybackslash\hspace{0pt}}p{#1}}

\begin{document}
\title{Practical Algorithms for STV and Ranked Pairs with Parallel Universes Tiebreaking}  
\author{Jun Wang, Sujoy Sikdar, Tyler Shepherd,  Zhibing Zhao, Chunheng Jiang, Lirong Xia}
\footnotetext{Rensselaer Polytechnic Institute, \{wangj38, sikdas, shepht2,zhaoz6,jiangc4\}@rpi.edu, xial@cs.rpi.edu}

\begin{abstract}
STV and ranked pairs (RP) are two well-studied voting rules for group decision-making. They proceed in multiple rounds, and are affected by how ties are broken in each round. However, the literature is surprisingly vague about how ties should be broken. We propose the first algorithms for computing the set of alternatives that are winners under {\em some} tiebreaking mechanism under STV and RP, which is also known as parallel-universes tiebreaking (PUT). Unfortunately, {\em PUT-winners} are NP-complete to compute under STV and RP, and standard search algorithms from AI do not apply. We propose multiple DFS-based algorithms along with pruning strategies and heuristics to prioritize search direction to significantly improve the performance using machine learning. We also propose novel ILP formulations for PUT-winners under STV and RP, respectively. Experiments on synthetic and real-world data show that our algorithms are overall significantly faster than ILP, while there are a few cases where ILP is significantly faster for RP. 
\end{abstract}


\maketitle

\section{Introduction}
The {\em Single Transferable Vote (STV)} rule\footnote{STV for choosing a winner is also known as {\em instant runoff voting, alternative vote}, or {\em ranked choice voting}.} is among the most popular voting rules used in real-world elections. According to Wikipedia, STV is being used to elect senators in Australia, city councils in San Francisco (CA, USA) and Cambridge (MA, USA), and more \cite{Wikipedia:STV}. In each round of STV, the lowest preferred alternative is eliminated, in the end leaving only one alternative, the winner, remaining.

This raises the question: {\em when two or more alternatives are tied for last place, how should we break ties to eliminate an alternative?} The literature provides no clear answer. For example, O'Neill lists many different STV tiebreaking variants ~\cite{stvvariants}. While the STV winner is unique and easy to compute for a fixed tiebreaking mechanism, it is NP-complete to compute all winners under {\em all} tiebreaking mechanisms. This way of defining winners is called parallel-universes tiebreaking (PUT) \cite{Conitzer09:Preference}, and we will therefore call them \textit{PUT-winners} in this paper.

Ties do actually occur in real-world votes under STV. On Preflib data ~\cite{Mattei13:Preflib}, 9.2\% of profiles have more than one PUT-winner under STV. There are two main motivations for computing all PUT-winners. First, it is vital in a democracy that the outcome not be decided by an arbitrary or random tiebreaking rule, which will violate the {\em neutrality} of the system ~\cite{Brill12:Price}. Second, even for the case of a unique PUT-winner, it is important to prove that the winner is unique despite ambiguity in tiebreaking. In an election, we would prefer the results to be transparent about who all the winners could have been. 

A similar problem occurs in the Ranked Pairs (RP) rule, which satisfies many desirable axiomatic properties in social choice~\cite{Schulze11:New}. The RP procedure considers every pair of alternatives and builds a ranking by selecting the pairs with largest victory margins. This continues until every pair is evaluated, the winner being the candidate which is ranked above all others by this procedure \cite{Tideman87:Independence}. Like in STV, ties can occur, and the order in which pairs are evaluated can result in different winners.  Unfortunately, like STV, it is NP-complete to compute all PUT-winners under RP~\cite{Brill12:Price}.

To the best of our knowledge, no algorithm beyond brute-force search is known for computing PUT-winners under STV and RP.  Given its importance as discussed above, the question we address in this paper is: {\em How can we design efficient, practical algorithms for computing PUT-winners under STV and RP?}

\subsection{Our Contributions}
Our main contributions are the first practical algorithms to compute the PUT-winners for STV and RP: search-based algorithms and integer linear programming (ILP) formulations. 

In our search-based algorithms, the nodes in the search tree represent intermediate steps in the STV and RP procedures, each leaf node is labeled with a single winner, and each root-to-leaf path represents a way to break ties. The goal is to output the union set of winners on the leaves. See Figure~\ref{fig:stvex} and Figure~\ref{fig:rpex} for examples. To improve the efficiency of the algorithms, we develop the following techniques: 

\noindent{\bf Pruning}, which maintains a set of {\em known PUT-winners} during the search procedure and can then prune a branch if expanding a state can never lead to any new PUT-winner.

\noindent{\bf Machine-learning-based prioritization}, which aims at building a large known winner set as soon as possible by prioritizing nodes that minimize the number of steps to discover a new PUT-winner. 

Our main conceptual contribution is a new measure called {\em early discovery}, wherein we time how long it takes to compute a given proportion of all PUT-winners on average. This is particularly important for low stakes and anytime applications, where we want to discover as many PUT-winners as possible with limited resources and at any point during execution. In addition, we design ILP formulations for STV and RP. 

The PUT problems are very challenging, mainly due to the exponential growth in the search space as the number of candidates increases (Section~\ref{sec:sizeofdata}). Yet our algorithms prove practical as experiments on synthetic and real-world data demonstrate. Specifically we show the following in the efficiency of our algorithms in solving the PUT problem for STV and RP, hereby denoted PUT-STV and PUT-RP respectively:\\
\noindent$\bullet$~For both PUT-STV and PUT-RP, in the large majority of cases our DFS-based algorithms are orders of magnitude faster than solving Integer Linear Programming (ILP) formulations in terms of total runtime and time to discover PUT-winners (Section~\ref{sec:ilp}).\\
\noindent$\bullet$~For PUT-STV, our devised priority function using machine learning results in significant reduction in time for discovering PUT-winners (Section~\ref{sec:stvheuristic}).\\
\noindent$\bullet$~For PUT-RP, \begin{enumerate*}[label=(\roman*)] \item our proposed pruning conditions exploit the structure of the RP procedure to provide a significant improvement in runtime (Section~\ref{sec:rpcp}), and \item our heuristic functions reduce both runtime and discovery time (Section~\ref{sec:rppriority}).\end{enumerate*}\\
\noindent$\bullet$~ Most hard profiles have two or more PUT-winners in synthetic datasets, while most real world profiles have single winner. Results show that running time increases with number of PUT-winners (Section~\ref{sec:dist}).\\

\section{Related Work and Discussions}\label{sec:related}
A previous version of this paper was presented at EXPLORE-17 workshop~\cite{Jiang:Practical}.
There is a large literature on the computational complexity of winner determination under commonly-studied voting rules. In particular, computing winners of the Kemeny rule has attracted much attention from researchers in AI and theory \cite{Conitzer06:Kemeny,Kenyon07:How}. However, STV and ranked pairs have both been overlooked in the literature, despite their popularity. We are not aware of previous work on practical algorithms for PUT-STV or PUT-RP. 
A recent work on computing winners of commonly-studied voting rules proved that computing STV is P-complete, but only with a fixed-order tiebreaking mechanism \cite{Csar2017:Winner}. Our paper focuses on finding all PUT-winners under all tiebreaking mechanisms.  See~\cite{Freeman2015:General} for more discussions on tiebreaking mechanisms in social choice.

Standard procedures to AI search problems unfortunately do not apply here. In a standard AI search problem, the goal is to find a path from the root to the goal state in the search space. However, for PUT problems, due to the unknown number of PUT-winners, we do not have a clear predetermined goal state. Other voting rules, such as Coombs and Baldwin, have similarly been found to be NP-complete to compute PUT winners~\cite{Mattei2014:How-hard}. The techniques we apply in this paper for STV and RP can be extended to these other rules, with slight modification based on details of the rule.

\section{Preliminaries}\label{sec:prelim}
Let $\ma=\{a_1,\cdots, a_m\}$ denote a set of $m$ {\em alternatives} and let $\ml(\ma)$ denote the set of all possible linear orders over $\ma$. A {\em profile} of $n$ voters is a collection $P =(V_i)_{i\le n}$ of votes where for each $i\leq n$, $V_i\in \ml(\ma)$. A voting rule takes as input a profile and outputs a non-empty set of winning alternatives. 

%

\noindent{\bf Single Transferable Vote (STV)}
proceeds in $m-1$ rounds over alternatives $\ma$ as follows. In each round, \begin{enumerate*}[label=(\arabic*)] \item an alternative with the lowest plurality score is eliminated, and \item the votes over the remaining alternatives are determined.\end{enumerate*} The last-remaining alternative is declared the winner.

\begin{figure}[H]
	\includegraphics[width=0.6\linewidth]{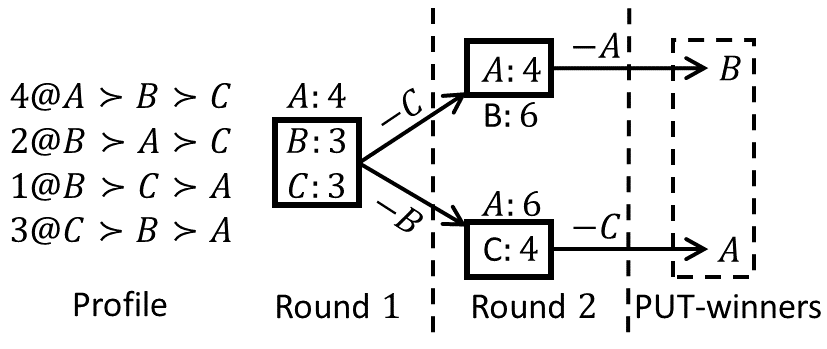}
	\caption{\label{fig:stvex} An example of the STV procedure.}
\end{figure}

\begin{ex}
	Figure~\ref{fig:stvex} shows an example of how the STV procedure can lead to different winners depending on the tiebreaking rule. In round 1, alternatives $B$ and $C$ are tied for last place. For any tiebreaking rule in which $C$ is eliminated, this leads to $B$ being declared the winner. Alternatively, if $B$ were to be eliminated, then $A$ is declared the winner.
\end{ex}

\noindent{\bf Ranked Pairs (RP)}. For a given profile $P=(V_i)_{i\le n}$, we define the 
{\em weighted majority graph} (WMG) of $P$, denoted by wmg$(P)$, to be the weighted digraph $(\ma,E)$ where the nodes are the alternatives, and for every pair of alternatives $a,b \in \ma$, there is an edge $(a,b)$ in $E$ with weight $w_{(a,b)} = |\{V_i: a \succ_{V_i} b\}| - |\{V_i: b \succ_{V_i} a\}|$. We define the \textit{nonnegative WMG} as wmg$_{\geq 0}(P) = (\ma, \{(a,b) \in E: w_{(a,b)} \ge 0\})$. We partition the edges of wmg$_{\geq 0}(P)$ into tiers $T_{1},...,T_{K}$ of edges, each with distinct edge weight values, and indexed according to decreasing value. Every edge in a tier $T_i$ has the same weight, and for any pair $i,j\le n$, if $i < j$, then $\forall e_1 \in T_i, e_2 \in T_j, w_{e_1} > w_{e_2}$.


\begin{figure}[h]
	\includegraphics[width=0.6\linewidth]{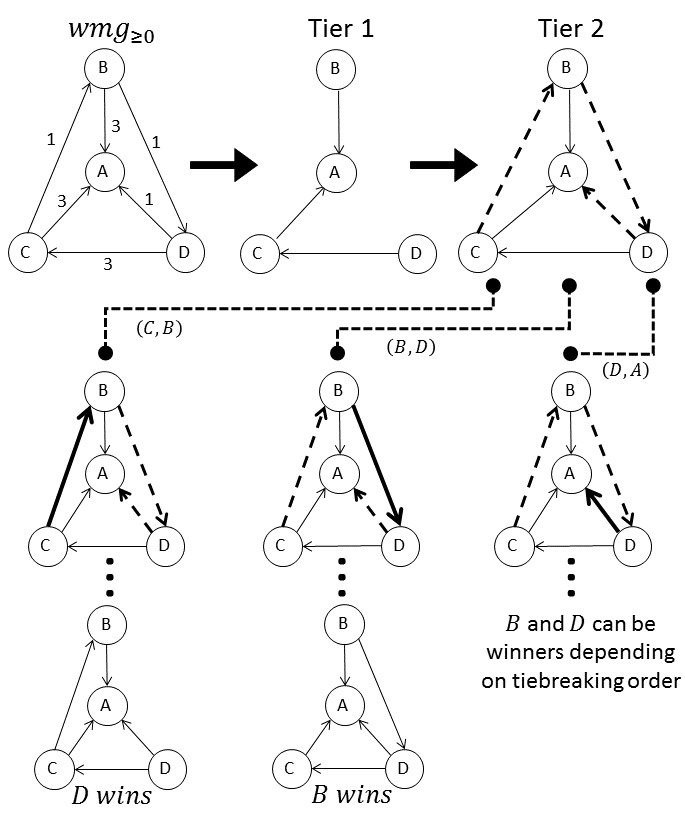}
	\caption{\label{fig:rpex} An example of the RP procedure.}
\end{figure}

Ranked pairs proceeds in $K$ rounds: Start with an empty graph $G$ whose vertices are $\mathcal A$. 
In each round $i \le K$, consider adding edges $e \in T_i$ to $G$ one by one according to a tiebreaking mechanism, as long as it does not introduce a cycle. 
Finally, output the ranking corresponding to the topological ordering of $G$, with the winner being ranked at the top.

\begin{ex}
	Figure~\ref{fig:rpex} shows the ranked pairs procedure applied to the WMG resulting from a profile over $m = 4$ alternatives (a profile with such a WMG always exists) \cite{McGarvey53:Theorem}. We focus on the addition of edges in tier $2$, where $\{(C,B) ,(B,D),(D,A)\}$ are to be added. Note that $D$ is the winner if $(C,B)$ is added first , while $B$ is the winner if $(B,D)$ is added first. 
\end{ex}

\section{Algorithms for PUT-STV}
We propose Algorithm~\ref{algo:stvdfs} to compute PUT-STV. For the most part, Algorithm~\ref{algo:stvdfs} follows a depth-first search (DFS) procedure, except that we include a \textit{pruning} condition whenever all alternatives remaining in the procedure are known to be winners, and the algorithm uses a heuristic \textit{priority} to order exploration of children. 


\begin{algorithm}[t]
	\SetAlgoNoLine
	\KwIn{A profile $P$.}
		\KwOut{All PUT-STV winners $W$.}
		Initialize a stack $F$ with the state $\ma$; $W=\emptyset$\;
		\While {$F$ is not empty}{
			Pop a state $S$ from $F$ to explore\;
			\If{$S$ has a single remaining alternative}{ add it to $W$\;}
			\If{$S$ already visited or $S \subseteq W$ (\textit{pruned})}{skip state\;}
			For every remaining lowest plurality score alternative $c$, in order of \textit{priority} add ($S \setminus c$) to $F$\;
		}
		\Return $W$\;
		\caption{PUT-STV$(P)$}
		\label{algo:stvdfs}
	\end{algorithm}

\noindent{\bf Early Discovery and Heuristic Function.} 
One advantage of Algorithm~\ref{algo:stvdfs} is its {\em any-time} property, which means that, if terminated at any time, it can output the known PUT-winners as an approximation to all PUT-winners. Such time constraint is realistic in low-stakes, everyday voting systems such as Pnyx~\cite{Brandt2015:Pnyx}, and it is desirable that an algorithm outputs as many PUT-winners as early as possible. To measure this, we introduce {\em early discovery} for PUT-winner algorithms. For any PUT-winner algorithm and any number $0\le \alpha\le 1$, the $\alpha$-discovery value is the average runtime for the algorithm to compute $\alpha$ fraction of PUT-winners. We note that $100\%$-discovery value can be much smaller than the total runtime of the algorithm, because the algorithm may continue exploring remaining nodes after $100\%$ discovery to verify that no new PUT-winner exists.

This is why we focus on DFS-based algorithms, as opposed to, for example, BFS---the former reaches leaf nodes faster. To achieve early discovery through Algorithm~\ref{algo:stvdfs}, we prioritize nodes whose state contains more candidate PUT-winners that have not been discovered. In this sense, we design a heuristic function for a state $S$ with known PUT-winners $W$, $Priority(S)=\sum_{c \in (S - W)} \pi(c)$. Here $\pi(c)$ is the machine learning model probability of $c$ to be a PUT-winner. Details of machine learning setup can be found in Section~\ref{sec:results}. It is important to note we do \textit{not} use the machine learning model to directly predict PUT-winners. 
Instead, in the searching process, if we are able to estimate the probability of a branch to have new PUT-winners, we can actively choose which branch to explore first. So under the circumstance without our knowing which branch is promising, machine learning can serve as our guidance to prioritize a better branch with higher probability to find PUT-winners.

\section{Algorithms for PUT-Ranked Pairs}\label{sec:rppractical}

At a high level, our algorithm takes a profile $P$ as input and solves PUT-RP using DFS. It is described as the $PUT$-$RP(P)$ procedure in Algorithm~\ref{algo:rpdfs}. Each node has a state $(G,E)$, where $E$ is the set of edges that have not been considered yet and $G$ is a graph whose edges are pairs that have been ``locked in'' by the RP procedure according to some tiebreaking mechanism. The root node is $(G = (\ma, \emptyset),E_0)$, where $E_0$ is the set of edges in wmg$_{\ge0}(P)$. 
Exploring a node $(G,E)$ at depth $t$ involves finding all maximal ways of adding edges from $T_t$ to $G$ without causing a cycle, which is done by the $MaxChildren()$ procedure shown in Algorithm~\ref{algo:maxchild}. $MaxChildren()$ takes a graph $G$ and a set of edges $T$ as input, and follows a DFS-like addition of edges one at a time. Within the algorithm, each node $(H,S)$ at depth $d$ corresponds to the addition of $d$ edges from $T$ to $H$ according to some tiebreaking mechanism. $S \subseteq T$ is the set of edges not considered yet.

\begin{dfn}
	Given a directed acyclic graph $G = (V,E)$, and a set of edges $T$, a graph $C=(V, E\cup T')$ where $T'\subseteq T$ is a {\em maximal child} of $(G,T)$ if and only if $\forall e\in T\backslash T'$, adding $e$ to the edges of $C$ creates a cyclic graph. 
\end{dfn}


\begin{algorithm}[t]
	\SetAlgoNoLine
	\KwIn{A profile $P$.} 
	\KwOut{All PUT-RP winners $W$.}
	Compute $(\ma,E_0) = \text{wmg}_{\ge 0}(P)$\;
	Initialize a stack $F$ with $((\ma,\emptyset),E_0)$ for DFS; $W = \emptyset$\;
	\While {$F$ is not empty}{
		Pop a state $(G, E)$ from $F$ to explore\;
		\If{$E$ is empty or this state can be pruned}{
			Add all topologically top vertices of $G$ to $W$ and skip state\;}
		$T \gets$ highest tier edges of $E$\;
		\For{$C$ in $MaxChildren(G,T)$}{
			Add $(C, E \setminus T)$ to $F$\;}
	}
	\Return $W$\;
	\caption{PUT-RP$(P)$}
	\label{algo:rpdfs}
\end{algorithm}

\begin{algorithm}[t]
	\SetAlgoNoLine
	\KwIn{A graph $G=(\ma,E)$, and a set of edges $T$.}
	\KwOut{Set $C$ of all maximal children of $G, T$.}
	Initialize a stack $I$ with $(G,T)$ for DFS; $C = \emptyset$\;
	\While{$I$ is not empty}{
		Pop $((\ma,E'),S)$ from $I$\;
		\If{$E'$ already visited or state can be \textit{pruned}}{ skip state\;}
		The successor states are $Q_e = (G_e,S\setminus{e})$ for each edge $e$ in $S$, where graph $G_e = (\ma, E' + e)$\;
		Discard states where $G_e$ is cyclic\;
		\eIf {in all successor states $G_e$ is cyclic}{We have found a max child; add $(\ma, E')$ to $C$\; }
		{Add states $Q_e$ to $I$ in order of \textit{local priority}\;}
	}
	\Return $C$\;
	\caption{MaxChildren$(G,T)$}
	\label{algo:maxchild}
\end{algorithm}


\noindent{\bf Pruning.}
For a graph $G$ and a tier of edges $T$, we implement the following conditions to check if we can terminate exploration of a branch of DFS early: \begin{enumerate*}[label=(\roman*)] \item If every alternative that is not a known winner has one or more incoming edges or \item If all but one vertices in $G$ have indegree $> 0$, the remaining alternative is a PUT-winner. \end{enumerate*} 
For example, in Figure~\ref{fig:rpex}, we can prune the right-most branch after having explored the two branches to its left.


\noindent{\bf Prioritization.} To aid in early discovery and faster pruning, we devised and tested three algorithms for heuristic functions. We will use $A$ to refer to the set of candidate PUT-winners (vertices with 0 indegree), and $W$ to refer to the set of known PUT-winners (previously discovered by the search).

\noindent$\bullet$~$\text{LP} = |A - W|$: Local priority; orders the exploration of children by the value of $|A - W|$, the number of potentially unknown PUT-winners.\\
\noindent$\bullet$~$\text{LPout} = \sum_{a \in A - W} \text{outdegree}(a)$: Local priority with outdegree.\\
\noindent$\bullet$~$\text{LPML} = \sum_{a \in A-W} \pi(a)$: Local priority with machine learning model $\pi$.

\begin{figure}
	\centering
	\includegraphics[scale = 0.7]{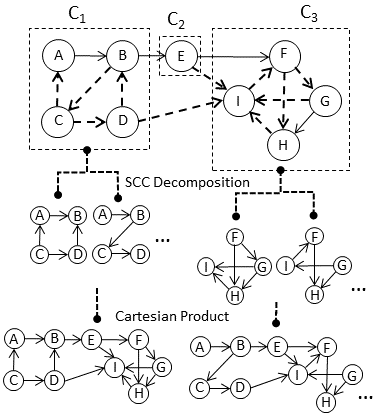}
	\caption{\label{signed graph_EP} Example of SCC Decomposition.} 
\end{figure}

\noindent{\bf SCC Decomposition.} We further improve Algorithm~\ref{algo:maxchild} by computing strongly connected components (SCCs). For a digraph, an SCC is a maximal subgraph of the digraph where for each ordered pair $u$, $v$ of vertices, there is a path from $u$ to $v$ . Every edge in an SCC is part of some cycle. The edges not in an SCC, therefore not part of any cycle, are called the bridge edges \cite[p.~98-99]{KleinbergEva:Algorithm}. Given a graph $G$ and a set of edges $T$, finding the maximal children will be simpler if we can split it into multiple SCCs. We find the maximal children of each SCC, then combine them in the Cartesian product with the maximal children of every other SCC. Finally, we add the bridge edges. Figure~\ref{signed graph_EP} shows an example of SCC Decomposition in which edges in $G$ are solid and edges in $T$ are dashed. Note this is only an example, and does not show all maximal children. In the unfortunate case when there is only one SCC we cannot apply SCC decomposition.

The following Theorem~\ref{thm:scc} is related to SCC decomposition used in solving the Feedback Arc Set problem, where finding minimal feedback arc sets is very similar to finding our maximal children. The minimal feedback arc set problem is to find, given a directed graph $G = (V,E)$, a minimal subset of edges $E' \subseteq E$ such that $G' = (V,E \setminus E')$ is acyclic. That is, adding any edge $e \in E'$ back to $G'$ creates a cycle \cite{baharev2015exact}. Our maximal children problem has the additional constraint that only edges in the tier $T$ can be removed from the edges of the graph.

\begin{thm}\label{thm:scc}
	For any directed graph $H$, $C$ is a maximal child of $H$ if and only if $C$ contains exactly \begin{enumerate*}[label=(\roman*)]
		\item all bridge edges of $H$ and \item the union of the maximal children of all SCCs in $H$.
	\end{enumerate*}
\end{thm}


\section{Experiment Results}\label{sec:results}

\subsection{Datasets}\label{sec:datasets}
We use both real-world preference profiles from Preflib and synthetic datasets with $m$ alternatives and $n$ voters to test our algorithms' performance. The synthetic datasets were generated based on impartial culture with $n$ independent and identically distributed rankings uniformly at random over $m$ alternatives for each profile.  From the randomly generated profiles, we only test on \textit{hard} cases where the algorithm encounters a tie that cannot be solved through simple pruning. All the following experiments are completed on an office machine with Intel i5-7400 CPU and 8GB of RAM running Python 3.5.

\begin{figure}[H]
	\centering
	\includegraphics[width=1\linewidth]{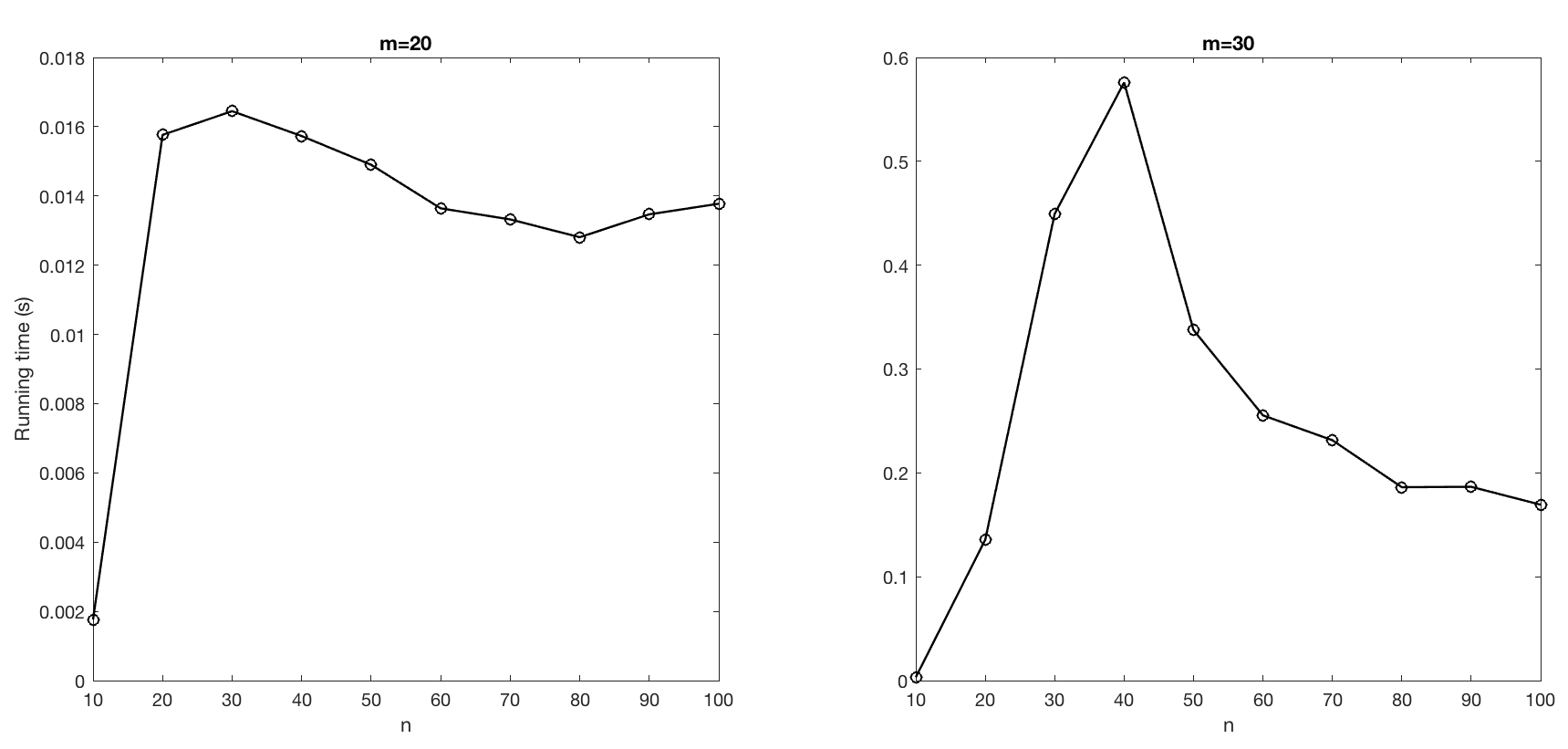}
	\caption{\label{P1P0} Running time of DFS for PUT-STV for different number of voters $n$, for profiles with $m=20$ and $m=30$ candidates.}
	
\end{figure}

\paragraph{Synthetic Data.} For PUT-STV, we used 50,000 $m=n=30$ synthetic profiles as our main dataset for the experiments in the paper. For PUT-RP, we generated 40,000 $m=n=10$ synthetic profiles at random, and picked out 14,875 hard profiles according to our definition in paragraph 1 of Section 6. In the method of SCC(LPML), as we stated in Section 6, we learned a neural network using tenfold cross validation on 10,000 hard profiles, and finally tested our algorithms on another 1,000 hard profiles. We chose the $m=n$ profiles as our dataset for both voting rules, because from Figure~\ref{P1P0} we can see that the running time reaches its peak when $m$ and $n$ are close. So in order to obtain relatively good performance on hard cases, we simply chose the $m=n$ profiles for testing.\\

\paragraph{Preflib Data.} For the real world data, we use all available datasets on Preflib suitable for our experiments on both rules. Specifically, 315 profiles from Strict Order-Complete Lists (SOC), and 275 profiles from Strict Order-Incomplete Lists (SOI). They represent several real world settings including political elections, movies and sports competitions. For political elections, the number of candidates is often not more than 30. For example, 76.1\% of 136 SOI election profiles on Preflib has no more than 10 candidates, and 98.5\% have no more than 30 candidates. 

\vspace{1em}

\subsection{PUT-STV}\label{sec:stvresults}
 We have the following observations.

\begin{figure}[]
	\centering
	\includegraphics[width=0.6\linewidth]{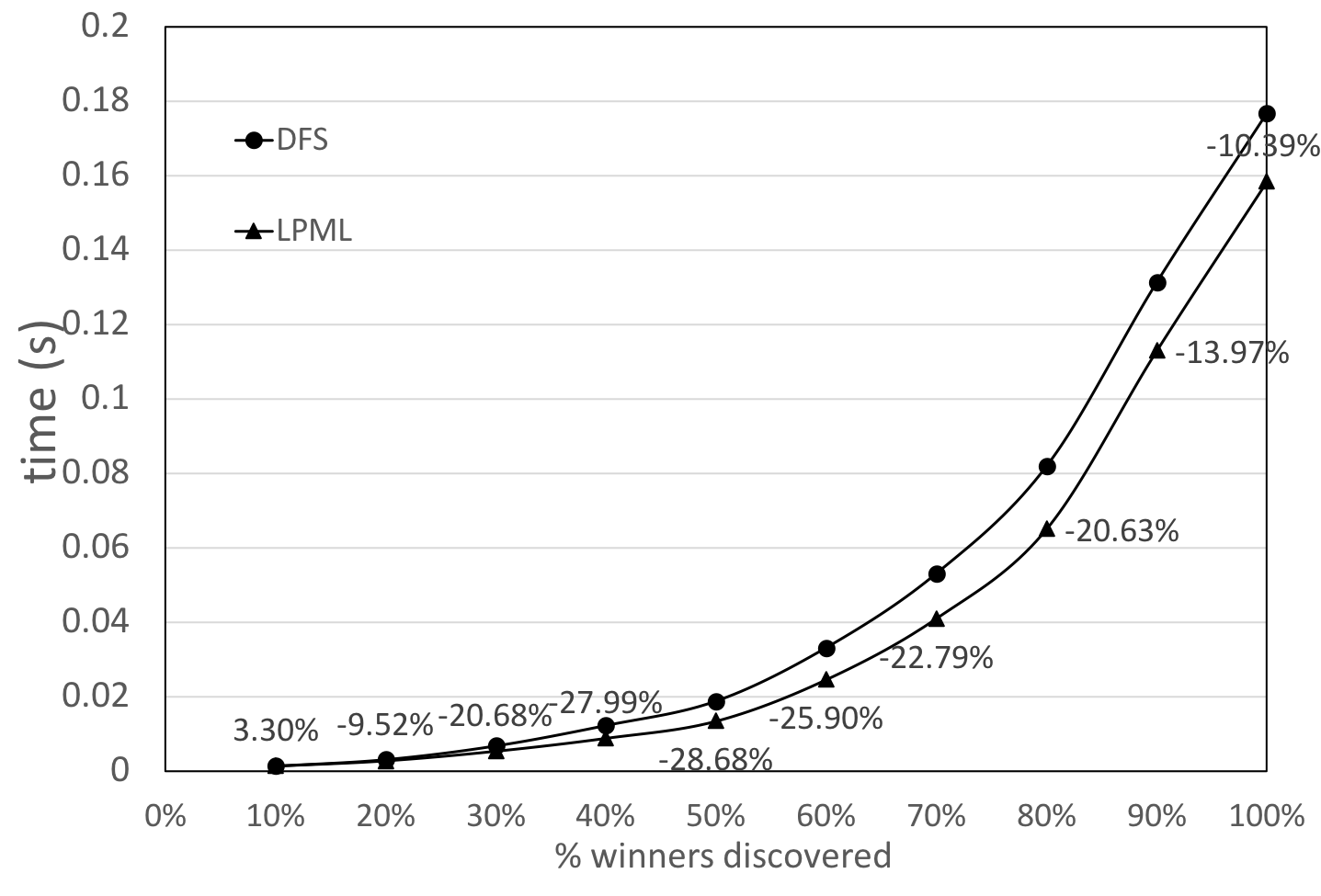}
	\caption{\label{fig:STV_discovery} PUT-STV early discovery.}
\end{figure}

\paragraph{Local Priority with Machine Learning Significantly Improves Early Discovery.}\label{sec:stvheuristic} As shown in Figure~\ref{fig:STV_discovery}, for $m=n=30$,  the algorithm of LPML which delivers a significant 10.39\% improvement over the baseline of an already optimal, manually designed DFS. \footnote{The baseline algorithm is already the best DFS algorithm without using machine learning, and is itself an important contribution of our work. } Further, the algorithm has $28.68\%$ reduction in 50\%-discovery. Results are similar for other datasets with different $m$. The early discovery figure is computed by averaging the time to compute a given percentage $p$ of PUT-winners. For example, for a profile with 2 PUT-winners which are discovered at time $t_1$ and $t_2$, we set the 10\%-50\% discovery time as $t_1$ and the 60\%-100\% discovery time as $t_2$.

For the machine learning in the local priority function, we learn a neural network model $\pi$ to predict the $m$-dimensional vector, where each component indicates whether the corresponding alternative is a  PUT-winner. We trained the models on 50,000 $m=n=30$ hard profiles using tenfold cross validation, with mean squared error $0.086$. We also tried other methods like SVC, kernel ridge regression and logistic regression.

\noindent{\em Pruning has Small Improvement.}\label{sec:stvcp} When evaluating the performance of pruning in PUT-STV, we see only a small improvement in the running time: on average, pruning brings only 0.33\% reduction in running time for $m=n=10$ profiles, 2.26\% for $m=n=20$ profiles, and 4.51\% for $m=n=30$ profiles.

\noindent{\em DFS is Practical on Real-World Data.} Our experimental results on Preflib data show that on 315 complete-order real world profiles, the maximum observed running time is only $0.06$ seconds and the average is 0.335 ms.


\begin{table*}
	\centering
	\begin{tabular}[width=\linewidth]{|c|c|c|c|c|c|} \hline
		&	DFS	&LP&SCC(LP)&  SCC(LP+outdeg)&	SCC(LPML)\\ \hline
		Avg. runtime (s)&	22.5045&	20.6086&	20.4445&	21.5017&	21.2949\\
		Avg. 100\%-discovery time (s) &	16.3914&	13.8476&	13.7339&	14.4154&	17.0183\\
		Avg. \# states & 52630.102 & 47967.451 & 47955.530 & 47959.605 & 48399.779\\
		Avg. \# prunes & 22434.844 & 20485.753 & 20474.685 & 20476.115 & 20655.721	\\
		\hline
	\end{tabular}
	\caption{\label{table:RP} Experiment results of different algorithms for finding maximal children in ranked pairs.}
\end{table*}

\vspace{1em}
\subsection{PUT-RP}\label{sec:rpresults} 
\begin{figure}[]
	\centering
	\includegraphics[width=0.7\linewidth]{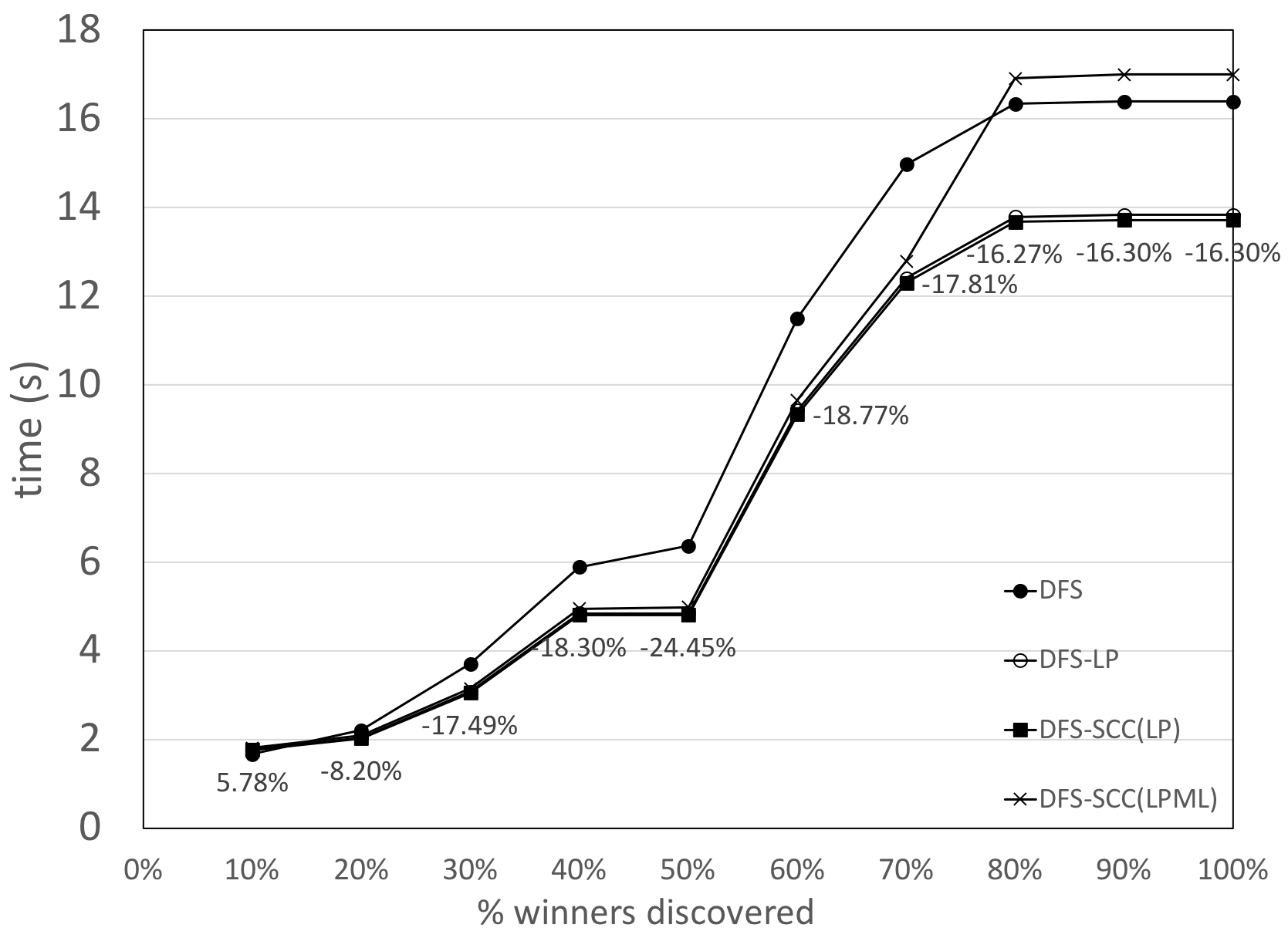}
	\caption{\label{fig:RP2} PUT-RP early discovery.} 
\end{figure}

We run different algorithms to find maximal children. Specifically, we evaluate four algorithms that use DFS with different improvements: \begin{enumerate*}[label=(\roman*)]\item standard DFS  (DFS in Figure~\ref{fig:RP2}), \item  local priority based on \# of candidate PUT-winners (LP), \item  local priority based on total outdegree of candidate PUT-winners (LPout),  and \item  local priority based on machine learning predictions (LPML).\end{enumerate*} We also evaluate the SCC based variants (denoted as SCC(x) where x is the original algorithm). Our experimental results are summarized in Table~\ref{table:RP}. We observe the following.


\paragraph{Pruning is Vital.}\label{sec:rpcp} Pruning plays a prominent part in the reduction of running time. From Table~\ref{table:RP}, we see that hitting the early stopping conditions always accounts for a large proportion (about 40\%) of the total number of states in the subfunction of finding maximal children. 
This means our pruning techniques avoid exploring many more states (than the number itself) under the eliminated branches. Our contrasting experiment further justifies this argument: on a dataset of 531 profiles, DFS without pruning takes 125.31 s in both running time and 100\%-discovery time on average, while DFS with pruning takes only 2.23 s and 2.18 s respectively with a surprising reduction of 98\%.

\paragraph{Local Priority Improves Performance.}\label{sec:rppriority} Our main conclusion is that SCC(LP) is the optimal algorithm for PUT-RP, as we see in Figure~\ref{fig:RP2}.  LP, i.e. local priority based on number of candidate PUT-winners, significantly reduces both average total running time and average time to discover all PUT-winners compared to standard DFS. SCC-based algorithms SCC(x) always perform slightly better than the corresponding algorithm x, due to the advantage in handling multi-SCC cases. In Figure~\ref{fig:RP2}, we compute the time-percentage relation for the 4 algorithms like in PUT-STV and plot the early discovery curves. Specifically, we show the reduction number of discovery time for SCC(LP) compared to DFS. We observe that SCC(LP) has the largest reduction in time; in particular it spends 24.45\%  less time compared to standard DFS when 50\% of PUT-winners are found. LP and SCC(LPout) are slightly worse, whereas SCC(LPML) does not help as much. For LPML,  we learn a neural network model $\pi$ using tenfold cross validation on a dataset of 10,000  $m=n=10$ profiles, with the objective of minimizing the $L1$-distance between the prediction vector and the target true winner vector. Our mean squared error was 0.0833 on a test set of 1,000 profiles.

\paragraph{Algorithms Perform Well on Real-World Data.} Using Preflib data, we find that our optimal algorithm SCC(LP) performs significantly better than standard DFS. We compare the two algorithms on 161 profiles with partial order. For SCC(LP), the average running time and 100\% discovery time are $1.33$s and $1.26$s, which have $46.0\%$ and $47.3\%$ reduction respectively compared to standard DFS. On 307 complete order profiles, the average running time and 100\% discovery time of SCC(LP) are both around 0.0179s with only a small reduction of 0.7\%, which is due to most profiles being easy cases without ties. In both experiments, we omit profiles with thousands of alternatives but very few votes which cause our machines to run out of memory.


\vspace{1em}
\subsection{Distribution of PUT-winners and Running Time}\label{sec:dist}

\begin{figure}[H]
	\centering
	\begin{tabular}{P{0.5\linewidth}P{0.5\linewidth}}
		\includegraphics[scale = 0.45]{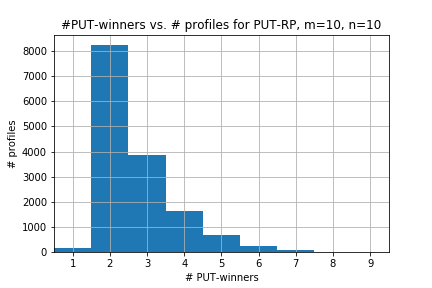}&\includegraphics[scale = 0.45]{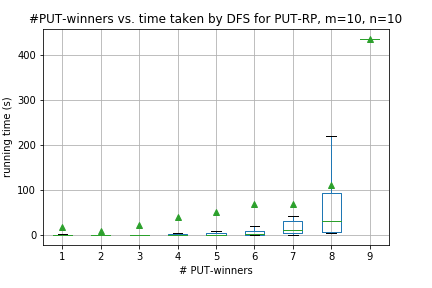} \\
		(a)  &	(b) 
	\end{tabular}
	\caption{\label{fig:rp10} Results for 14,875 hard profiles for PUT-RP with $m=10, n= 10$. (a) shows the histogram of \# PUT-winners in hard synthetic datasets for PUT-RP with $m=10, n= 10$. (b) shows the running time of DFS for PUT-RP vs. \# PUT-winners for profiles with $m=10, n= 10$. Green arrows show the mean. Green horizontal line shows the median (second quartile). Box shows first and third quartiles. Whiskers show minimum and maximum.} 
	
\end{figure}

\begin{figure}[htp]
	\centering
	\begin{tabular}{P{0.5\linewidth}P{0.5\linewidth}}
		\includegraphics[scale = 0.45]{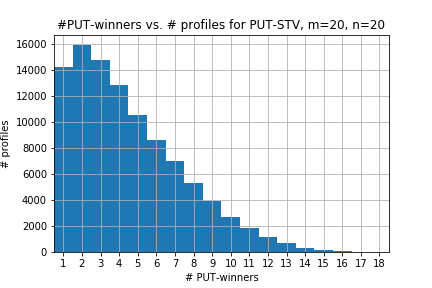} & \includegraphics[scale = 0.45]{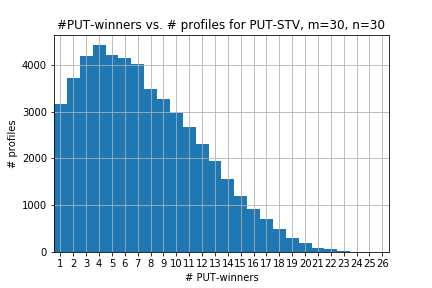}\\
		(a) & (c) \\
		\includegraphics[scale = 0.45]{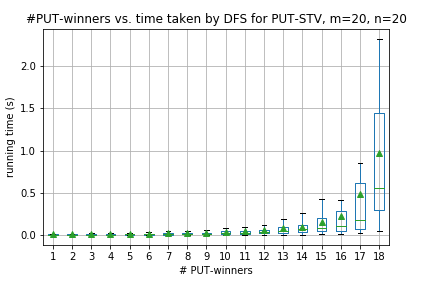} & \includegraphics[scale = 0.45]{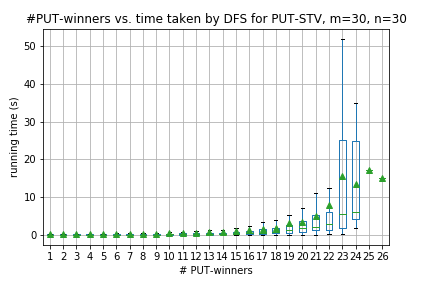}\\
		(b) & (d)
	\end{tabular}
	\caption{\label{fig:stv} Results for PUT-STV with 100,000 hard synthetic profiles for $m=20, n= 20$, and 50,000 hard synthetic profiles for $m=30, n=30$. (a) and (c) show the histogram of number of PUT-winners for $m=20, n= 20$, and $m=30, n=30$ respectively. (b) and (d) show the running time of DFS for PUT-STV vs. number of PUT-winners for profiles with $m=20, n= 20$, and $m=30, n=30$ respectively. Green arrows show the mean. Green horizontal line shows the median (second quartile). Box shows first and third quartiles. Whiskers show minimum and maximum values.} 
	
\end{figure}

\paragraph{Majority of hard profiles have two or more PUT-winners in synthetic datasets.}
As we show in Figure~\ref{fig:rp10}(a), for PUT-RP with $m=10, n=10$, $>99\%$ of the 14,875 hard synthetic profiles have two or more PUT-winners. Similarly, Figures~\ref{fig:stv}(a), and ~\ref{fig:stv}(c) show the histogram of the number of PUT-winners in all synthetic profiles used in our experiments for PUT-STV with $m=20,n=20$, and PUT-STV with $m=30,n=30$ respectively. We find that greater than two-thirds of the profiles have two or more PUT-winners in these experiments. 

\paragraph{Most real world profiles have single winner.} 90.8\% out of 315 SOC profiles have single winner under PUT-STV; 93.2\% out of 307 non-timeout SOC profiles, and 89.4\% out of 161 non-timeout SOI profiles have single PUT-RP winner. 

\paragraph{Running time increases with number of PUT-winners.}
As we show in Figure~\ref{fig:rp10}(b), for PUT-RP with $m=10, n=10$, the running time grows with the number of PUT-winners. We make the same observation for PUT-STV (see Figure~\ref{fig:stv}(b) for $m=20, n=20$, and Figure~\ref{fig:stv}(d) for $m=30, n=30$). 

\vspace{1em}
\subsection{The impact of the size of datasets on the algorithms}\label{sec:sizeofdata} 

The sizes of $m$ and $n$ have different effects on searching space. Our algorithms can  deal with larger numbers of voters ($n$) without any problem. In fact, increasing $n$ reduces the likelihood of ties, which makes the computation easier. 

But for larger $m$,  the issue of memory constraint which comes from using cache to store visited states, becomes crucial. Without using cache, DFS becomes orders of magnitude slower. Our algorithm for PUT-STV with $m>30$ terminates with memory errors due to the exponential growth in state space, and our algorithm for PUT-RP is in a similar situation. Even with as few as $m = 10$ alternatives, the search space grows large. There are $ 3 ^ {\binom{m}{2}} = 3 ^ {\frac{m(m-1)}{2}} $ possible states of the graph. For $m = 10$, this is $2.95\times10^{21}$ states. As such, due to memory constraints, currently we are only able to run our algorithms on profiles of size $m = n = 10$ for PUT-RP.

\section{Integer Linear Programming}\label{sec:ilp}

\noindent{\bf ILP for PUT-STV and Results.} The solutions correspond to the elimination of a single alternative in each of $m-1$ rounds and we test whether a given alternative is the PUT-winner by checking if there is a feasible solution when we enforce the constraint that the given alternative is not eliminated in any of the rounds. We omit the details due to the space constraint. Table~\ref{tbl:ilp_stv} summarizes the experimental results obtained using Gurobi's ILP solver. Clearly, the ILP solver takes far more time than even our most basic search algorithms without improvements.

\begin{table}[htp]
	\centering	
	\begin{tabular}[width=\linewidth]{|r|c|c|c|}\hline
		$m$& 10 & 20 & 30 \\
		$n$ & 10 & 20 & 30\\ \hline
		\# Profiles & 1000 & 2363 & 21\\
		Avg.~Runtime(s) & 1.14155 & 155.1874 & 12877.2792\\ \hline
	\end{tabular}
	\caption{\label{tbl:ilp_stv}ILP for STV rule.}
\end{table}


\noindent{\bf ILP for PUT-RP.} We develop a novel ILP based on the characterization by Zavist and Tideman (Theorem~\ref{thm:iw}).
Let the {\em induced weight} (IW) between two vertices $a$ and $b$ be the maximum path weight over all paths from $a$ to $b$ in the graph. The path weight is defined as the minimum edge weight of a given path. An edge $(u,v)$ is {\em consistent} with a ranking $R$ if $u$ is preferred to $v$ by $R$. $G_R$ is a graph whose vertices are $\ma$ and whose edges are exactly every edge in $\text{wmg}_{\geq0}(P)$ consistent with a ranking $R$. Thus there is a topological ordering of $G_R$ that is exactly $R$.


\begin{ex} In Figure~\ref{fig:rpex}, consider the induced weight from $D$ to $A$ in the bottom left graph. There are three distinct paths: $P_1 = \{D \rightarrow A\}$, $P_2 = \{D \rightarrow C \rightarrow A\}$, and $P_3 = \{D \rightarrow C \rightarrow B \rightarrow A\}$. The weight of $P_1$, or $W(P_1) = 1$, $W(P_2) = 3$ and $W(P_3) = 1$. Thus, IW$(D,A) = 3$, and note that IW$(D,A) \geq w_{(A,D)} = -1$.
\end{ex}

\begin{thm}~\cite{zavist1989} \label{thm:iw}
	For any profile $P$ and for any strict ranking $R$, the ranking $R$ is the outcome of the ranked pairs procedure if and only if $G_R$ satisfies the following property for all candidates $i,j \in \ma$: $\forall i \succ_R j, \text{ IW}(i,j) \geq w_{(j,i)}$.
\end{thm}


Based on Theorem~\ref{thm:iw}, we provide a novel ILP formulation of the PUT-RP problem. We can test whether a given alternative $i^*$ is a PUT-RP winner if there is a solution subject to the constraint that there is no path from any other alternative to $i^*$. The variables are: \begin{enumerate*}[label=(\roman*)]
	\item A binary indicator variable $X_{i,j}^t$ of whether there is an $i\to j$ path using locked in edges from $\bigcup T_{i\le t}$, for each $i,j\le m, t\le K$.
	\item A binary indicator variable $Y_{i,j,k}^t$ of whether there is an $i\to k$ path involving node $j$ using locked in edges from tiers $\bigcup T_{i\le t}$, for each $i,j,k\le m, t\le K$.
\end{enumerate*}

We can determine all PUT-winners by selecting every alternative $i^* \leq m$, adding the constraint $\sum_{j\le m, j\neq i^*} X_{j,i^*}^K = 0$, and checking the feasibility with the following constraints:\\
\noindent$\bullet$~To enforce Theorem~\ref{thm:iw}, for every pair $i, j \le m$, such that $(j,i) \in T_t$, we add the constraint $X_{i,j}^{t} \ge X_{i,j}^K$.\\
\noindent$\bullet$~In addition, we have constraints to ensure that \begin{enumerate*}[label=(\roman*)]\item locked in edges from $\bigcup_{t\le K} T_t$ induce a total order over $\ma$ by enforcing asymmetry and transitivity constraints on $X_{i,j}^K$ variables, and \item enforcing that if $X_{i,j}^t=1$, then $X_{i,j}^{\hat t >t}=1$.\end{enumerate*}\\
\noindent$\bullet$~The constraints to ensure that maximum weight paths are selected are detailed in Figure~\ref{fig:ilp}.
\begin{figure}[H]
	\centering
	\begin{align*}
	\left.~~~\begin{tabular}{l}
	$\forall i,j,k\le m, t\le K$,\\
	$Y_{i,j,k}^t \ge X_{i,j}^t + X_{j,k}^t - 1$\\
	$Y_{i,j,k}^t \le\frac{X_{i,j}^t + X_{j,k}^t}{2}$
	\end{tabular}\right\}& \text{$i\to j \to k$}
	\end{align*}
	\vspace{1em}
	\begin{align*}
	\left.\begin{tabular}{l}
	$\forall i,k\le m, t \le K,$\\ 
	if $(i,k) \in E^{\hat t \le t}$, $X_{i,k}^t \ge X_{i,k}^K$
	\end{tabular}\right\}& \text{$(i,k)$}\\
	\end{align*}
	\begin{align*}
	\left.\begin{tabular}{P{2.4cm}l}
	& $\forall j \le m$, \\
	& $X_{i,k}^t \ge Y_{i,j,k}^t$,\\
	if $(i,k) \in T_{\hat t > t}$, & $X_{i,k}^t\le \sum_{j\le m}Y_{i,j,k}^t$,\\
	if $(i,k) \in T_{\hat t \le t}$, & $X_{i,k}^t\le \sum_{j\le m}Y_{i,j,k}^t + X_{i,k}^K$\\
	\end{tabular}\right\} & \text{$i\to k$}\\
	\end{align*}
	\caption{Maximum weight path constraints.} \label{fig:ilp}
\end{figure}

\noindent{\bf Results.} Out of $1000$ hard profiles, the RP ILP ran faster than DFS on $16$ profiles. On these $16$ profiles, the ILP took only $41.097\%$ of the time of the DFS to compute all PUT-winners on average. However over all $1000$ hard profiles, DFS is significantly faster on average: 29.131 times faster. We propose that on profiles where DFS fails to compute all PUT-winners, or for elections with a large number of candidates, we can fall back on the ILP to solve PUT-RP.
\section{Future Work}\label{sec:cls}

There are many other strategies we wish to explore. In the local priority method, we implemented multiple priority functions, but none of them are significantly better than the number of potential PUT-winners. So one future work is to find a better priority function to encourage early discovery of new winners. Further machine learning techniques or potentially reinforcement learning could prove useful here. For PUT-RP, we want to specifically test the performance of our SCC-based algorithm on large profiles with many SCCs, since currently our dataset contains a low proportion of multi-SCC profiles. Also, we want to extend our search algorithm  to multi-winner voting rules like the Chamberlin–Courant rule, which is known to be NP-hard to compute an optimal committee for general preferences \cite{Procaccia:Multi}.

	
	



\bibliographystyle{named}
\bibliography{../references}

\begin{thebibliography}{}

\bibitem[\protect\citeauthoryear{Baharev \bgroup \em et al.\egroup
  }{2015}]{baharev2015exact}
Ali Baharev, Hermann Schichl, Arnold Neumaier, and TOBIAS Achterberg.
\newblock An exact method for the minimum feedback arc set problem.
\newblock {\em University of Vienna}, 10:35--60, 2015.

\bibitem[\protect\citeauthoryear{Brandt and Geist}{2015}]{Brandt2015:Pnyx}
Felix Brandt and Guillaume Chabinand~Christian Geist.
\newblock {Pnyx: A Powerful and User-friendly Tool for Preference Aggregation}.
\newblock In {\em Proceedings of the 2015 International Conference on
  Autonomous Agents and Multiagent Systems}, pages 1915--1916, 2015.

\bibitem[\protect\citeauthoryear{Brill and Fischer}{2012}]{Brill12:Price}
Markus Brill and Felix Fischer.
\newblock {The Price of Neutrality for the Ranked Pairs Method}.
\newblock In {\em Proceedings of the National Conference on Artificial
  Intelligence (AAAI)}, pages 1299--1305, Toronto, Canada, 2012.

\bibitem[\protect\citeauthoryear{Conitzer \bgroup \em et al.\egroup
  }{2006}]{Conitzer06:Kemeny}
Vincent Conitzer, Andrew Davenport, and Jayant Kalagnanam.
\newblock Improved bounds for computing {K}emeny rankings.
\newblock In {\em Proceedings of the National Conference on Artificial
  Intelligence (AAAI)}, pages 620--626, Boston, MA, USA, 2006.

\bibitem[\protect\citeauthoryear{Conitzer \bgroup \em et al.\egroup
  }{2009}]{Conitzer09:Preference}
Vincent Conitzer, Matthew Rognlie, and Lirong Xia.
\newblock Preference functions that score rankings and maximum likelihood
  estimation.
\newblock In {\em Proceedings of the Twenty-First International Joint
  Conference on Artificial Intelligence (IJCAI)}, pages 109--115, Pasadena, CA,
  USA, 2009.

\bibitem[\protect\citeauthoryear{Csar \bgroup \em et al.\egroup
  }{2017}]{Csar2017:Winner}
Theresa Csar, Martin Lackner, Reinhard Pichler, and Emanuel Sallinger.
\newblock {Winner Determination in Huge Elections with MapReduce}.
\newblock In {\em Proceedings of the AAAI Conference on Artificial
  Intelligence}, 2017.

\bibitem[\protect\citeauthoryear{Freeman \bgroup \em et al.\egroup
  }{2015}]{Freeman2015:General}
Rupert Freeman, Markus Brill, and Vincent Conitzer.
\newblock {General Tiebreaking Schemes for Computational Social Choice}.
\newblock In {\em Proceedings of the 2015 International Conference on
  Autonomous Agents and Multiagent Systems}, pages 1401--1409, 2015.

\bibitem[\protect\citeauthoryear{Jiang \bgroup \em et al.\egroup
  }{2017}]{Jiang:Practical}
Chunheng Jiang, Sujoy Sikdar, Jun Wang, Lirong Xia, and Zhibing Zhao.
\newblock Practical algorithms for computing stv and other multi-round voting
  rules.
\newblock In {\em EXPLORE-2017: The 4th Workshop on Exploring Beyond the Worst
  Case in Computational Social Choice}, 2017.

\bibitem[\protect\citeauthoryear{Kenyon-Mathieu and
  Schudy}{2007}]{Kenyon07:How}
Claire Kenyon-Mathieu and Warren Schudy.
\newblock {How to Rank with Few Errors: A PTAS for Weighted Feedback Arc Set on
  Tournaments}.
\newblock In {\em Proceedings of the Thirty-ninth Annual ACM Symposium on
  Theory of Computing}, pages 95--103, San Diego, California, USA, 2007.

\bibitem[\protect\citeauthoryear{Kleinberg and
  Tardos}{2005}]{KleinbergEva:Algorithm}
Jon Kleinberg and Eva Tardos.
\newblock {\em Algorithm Design}.
\newblock Pearson, 2005.

\bibitem[\protect\citeauthoryear{Mattei and Walsh}{2013}]{Mattei13:Preflib}
Nicholas Mattei and Toby Walsh.
\newblock {PrefLib: A Library of Preference Data}.
\newblock In {\em {Proceedings of Third International Conference on Algorithmic
  Decision Theory (ADT 2013)}}, {Lecture Notes in Artificial Intelligence},
  2013.

\bibitem[\protect\citeauthoryear{Mattei \bgroup \em et al.\egroup
  }{2014}]{Mattei2014:How-hard}
Nicholas Mattei, Nina Narodytska, and Toby Walsh.
\newblock How hard is it to control an election by breaking ties?
\newblock In {\em Proceedings of the Twenty-first European Conference on
  Artificial Intelligence}, pages 1067--1068, 2014.

\bibitem[\protect\citeauthoryear{McGarvey}{1953}]{McGarvey53:Theorem}
David~C. McGarvey.
\newblock A theorem on the construction of voting paradoxes.
\newblock {\em Econometrica}, 21(4):608--610, 1953.

\bibitem[\protect\citeauthoryear{O'Neill}{2011}]{stvvariants}
Jeff O'Neill.
\newblock \url{https://www.opavote.com/methods/single-transferable-vote}, 2011.

\bibitem[\protect\citeauthoryear{Procaccia \bgroup \em et al.\egroup
  }{2007}]{Procaccia:Multi}
Ariel~D Procaccia, Jeffrey~S Rosenschein, and Aviv Zohar.
\newblock Multi-winner elections: Complexity of manipulation, control and
  winner-determination.
\newblock In {\em IJCAI}, volume~7, pages 1476--1481, 2007.

\bibitem[\protect\citeauthoryear{Schulze}{2011}]{Schulze11:New}
Markus Schulze.
\newblock A new monotonic, clone-independent, reversal symmetric, and
  {C}ondorcet-consistent single-winner election method.
\newblock {\em Social Choice and Welfare}, 36(2):267---303, 2011.

\bibitem[\protect\citeauthoryear{Tideman}{1987}]{Tideman87:Independence}
T.~Nicolaus Tideman.
\newblock Independence of clones as a criterion for voting rules.
\newblock {\em Social Choice and Welfare}, 4(3):185--206, 1987.

\bibitem[\protect\citeauthoryear{Wikipedia}{2018}]{Wikipedia:STV}
Wikipedia.
\newblock Single transferable vote --- {W}ikipedia{,} the free encyclopedia,
  2018.
\newblock [Online; accessed 30-Jan-2018].

\bibitem[\protect\citeauthoryear{Zavist and Tideman}{1989}]{zavist1989}
T.~M. Zavist and T.~N. Tideman.
\newblock Complete independence of clones in the ranked pairs rule.
\newblock {\em Social Choice and Welfare}, 6(2):167--173, Apr 1989.

\end{thebibliography}

\end{document}